\documentclass[journal=jacsat,manuscript=article]{achemso}

\usepackage{chemformula} 
\usepackage[T1]{fontenc} 
\usepackage{xcolor}
\usepackage{subcaption}

\usepackage{afterpage}

\author{Connor Robertson}
\affiliation{Department of Mathematics, New Jersey Institute of Technology, Newark, NJ, USA}
\author{Jared L. Wilmoth}
\affiliation{Department of Environmental Science and Technology, University of Maryland, College Park, MD 20742, USA
}
\author{Scott Retterer}
\author{Miguel Fuentes-Cabrera}
\affiliation{Center for Nanophase Materials Sciences, Oak Ridge National Laboratory, Oak Ridge, TN, 37831, USA}
\email{fuentescabma@ornl.gov}

\title{Performing Video Frame Prediction of Microbial Growth with a Recurrent Neural Network} 


\begin{document}



\vspace*{\fill} 
\begin{center}
\fbox{
    \begin{minipage}{.75\textwidth}
    \linespread{1.0}\selectfont
        Notice: This manuscript has been authored by UT-Battelle, LLC under  Contract No. DE-AC05-00OR22725 with the U.S. Department of Energy.  The  United States Government retains and the publisher, by accepting the  article for publication, acknowledges that the United States Government  retains a non-exclusive, paid-up, irrevocable, world-wide license to publish  or reproduce the published form of this manuscript, or allow others to do  so, for United States Government purposes.  The Department of Energy will  provide public access to these results of federally sponsored research in  accordance  with  the  DOE  Public  Access  Plan  (\url{http://energy.gov/downloads/doe-public-access-plan}). 
    \end{minipage}
}
\end{center}
\vspace*{\fill}

\clearpage

\maketitle

\begin{abstract}
 A Recurrent Neural Network (RNN) was used to perform video frame prediction of microbial growth for a population of two mutants of {\it Pseudomonas aeruginosa}. The RNN was trained on videos of 20 frames that were acquired using fluorescence microscopy and microfluidics. The network predicted the last 10 frames of each video, and the accuracy's of the predictions was assessed by comparing raw images, population curves, and the number and size of individual colonies. Overall, we found the predictions to be accurate using this approach. The implications this result has on  designing autonomous experiments in microbiology, and the steps that can be taken to make the predictions even more accurate, are  discussed.
 \end{abstract}

\section{Introduction}
Recurrent Neural Networks, RNNs, are a type of Artificial Neural Network that takes temporal sequences as inputs. RNNs were originally created for tasks related to Natural Language Processing, such as text classification and translation. The addition of convolutions extended the use of RNNs to data consisting of temporal sequences of images ({\it i.e.} videos). RNNs are currently used for a variety of image-based tasks, which include  video captioning and video frame prediction.

Over the last few years, new advances have  made RNNs' video frame prediction more accurate; a comprehensive review on this matter can be found in \cite{9294028}. One of such advancements is known as  PredRNN\cite{DBLP:journals/corr/abs-2103-09504}, which has been used for performing video frame prediction for videos of numbers moving across a screen (moving MNIST database), humans performing different actions (KTH database), traffic flow, and moisture movement in weather systems\cite{DBLP:journals/corr/abs-2103-09504}. In each of these cases, PredRNN \cite{DBLP:journals/corr/abs-2103-09504} outperformed other RNNs.
Here we investigated whether PredRNN could be used to perform video frame prediction in videos of microbial growth.

Training an RNN to  perform  video frame prediction requires a large number of  videos. For example, for predicting the movement of numbers across a screen with PredRNN,  10000 videos of 20 frames were needed \cite{DBLP:journals/corr/abs-2103-09504}. 
Unfortunately, experimental image acquisition techniques rarely produce such a large quantity of videos. 
Fortunately, in the case of microbiology, there's a technique that has the potential of producing larger amounts of videos. This technique leverages the combination of microfluidics and fluorescence microscopy in a high-throughput experimental platform.

\begin{figure}[H]
  \includegraphics[width=\textwidth]{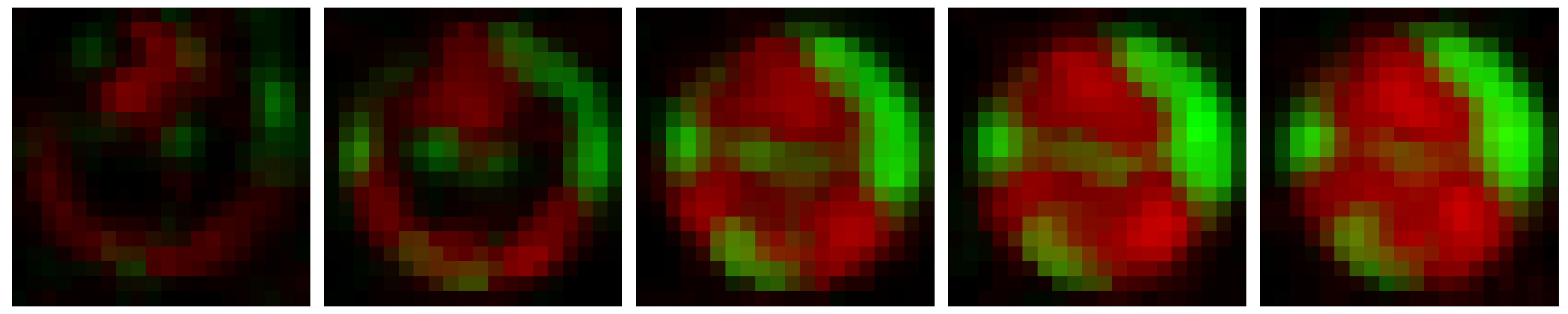}
  \caption{ Brightened snapshots at time steps 1, 4, 7, 10, 13, for one of the videos obtained in \cite{WOS:000456223900011} for the 30$\mu$m well.}
  \label{fig:fig2}
\end{figure}

Timm \textit{et al.} used microfluidics and fluorescence microscopy to investigate the growth of \textit{Pseudomonas aeruginosa}, an antiobiotic resistant bacteria  responsible for many clinical infections\cite{WOS:000456223900011}. Using a  photolithographic procedure, Timm \textit{et al.} built a chip consisting of an array of microwells with wells of different  diameters~\cite{WOS:000407448100034}. The diameters ranged from 5 to 100$\mu$m, and each well had several replicas to ensure the data collected was statistically significant.
 The chip was seeded with a microbial population that contained two mutant strains of \textit{P. aeruginosa} at different  concentrations\cite{WOS:000407448100034}. One mutant possessed the Type VI Secretion System, T6SS, while the other lacked it. The T6SS involves a needle-like pilus that is used  to attack other bacteria\cite{WOS:000404511700002}. 
 Both  mutants were susceptible to T6SS attacks. To monitor the growth of this population,  the T6SS-mutant and non-T6SS-mutant  were  tagged with green and red fluorescent protein, respectively,  and their growth monitored with fluorescence microscopy.
This microwell fabrication and seeding process has opened the door to larger scale data-driven analysis of bacterial population interactions~\cite{10.1371/journal.pone.0155080}\cite{doi:10.1116/1.4962739}.
Five snapshots of one of the microwell videos obtained by  Timm \textit{et al.}\cite{WOS:000456223900011} are shown in Fig.\ref{fig:fig2}.

Here we show that PredRNN \cite{DBLP:journals/corr/abs-2103-09504} can be used to accurately predict future frames of videos such as those shown in Fig.\ref{fig:fig2}. The accuracy of the predictions is evaluated with common image-based neural network metrics, such as those used in Ref.\cite{DBLP:journals/corr/abs-2103-09504}, as well as with metrics that measured total population curves and number and size of individual colonies. We believe that the results presented here pave the way for designing autonomous experiments in microbiology by using PredRNN \cite{DBLP:journals/corr/abs-2103-09504} in conjunction with edge computing devices. 

\section{Results and Discussion}
Timm {\it et. al}\cite{WOS:000407448100034} collected fluorescence microscopy images every 30 minutes for a period of 24 hours. Of all the well sizes investigated in \cite{WOS:000407448100034}, we selected the 30$\mu$m wells as the most useful due to the observation that growing microbial biomass completely filled these wells during the laboratory study and also led to distinct partitioning of the co-cultured mutant cell types due to T6SS interactions (influenced by spatial confinement in these wells).
Thus, we focused the present study on images collected using the 30$\mu$m wells.

A total of 48 videos of 14 frames were collected for the  30$\mu$m well size, with each video showing microbial growth in a single well. 
Within each video, frames 1 to 7 showed the most interesting dynamics, which corresponded to the exponential growth of the population  from a few initial colonies to a large community that covered the entirely  of the well.
By comparison, frames 8 to 14 were practically static, having reached a saturation state around frame 8.
(After frame 14, the population begins to decrease due to the lack of nutrients, but this stage is not considered in this study.)
For video frame prediction, frames 1 to 7 present a nontrivial challenge, and in what follows we considered these frames only.
Effectively, this gave us 48 videos of 7 frames.

\begin{figure}[H]
     \begin{subfigure}{.32\textwidth}
       \includegraphics[width=\textwidth]{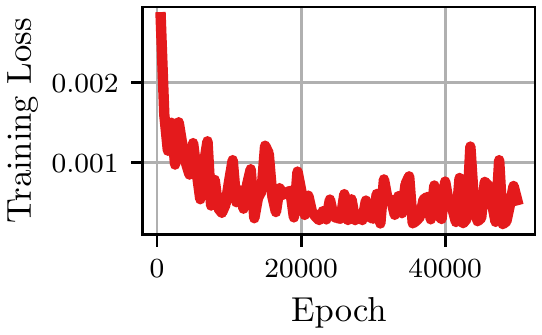}
       \subcaption{}
     \end{subfigure}
     \begin{subfigure}{.32\textwidth}
       \includegraphics[width=\textwidth]{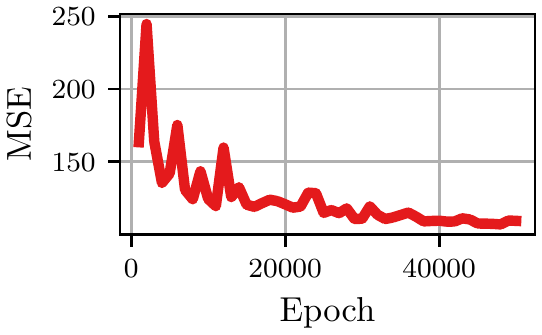}
       \subcaption{}
     \end{subfigure}
     \begin{subfigure}{.32\textwidth}
       \includegraphics[width=\textwidth]{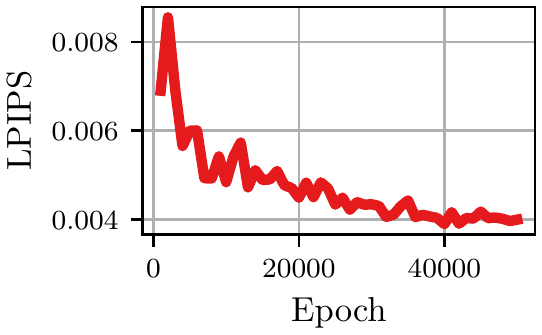}
       \subcaption{}
     \end{subfigure}
   \caption{Convergence metrics of PredRNN\cite{DBLP:journals/corr/abs-2103-09504} training. (a) MSE loss for training images. (b) MSE loss for validation images. (c) LPIPS loss for validation images.}
   \label{training_loss}
 \end{figure}

Temporal and spatial transformations, described in section {\it Methods}, were used to augment the number of frames in each video, after which a group of test wells was separated and later used to assess the generalization and transfer ability of the trained network.
After more data augmentation to increase the total number of videos, 392 videos of 20 frames were obtained.
The videos were subsequently separated into two datasets, training and validation, in a proportion of 80\% and 20\%, which were used to train PredRNN \cite{DBLP:journals/corr/abs-2103-09504}.
During training, PredRNN\cite{DBLP:journals/corr/abs-2103-09504} learns the spatiotemporal dynamics of the videos, and this is used to make predictions.
Specifically, of the 20 frames that make up each video, the first 10 were used as input and the remaining ones were used for prediction.
In what follows, the original 11-20 frames  will be referred to as groundtruth, while the corresponding frames predicted by PredRNN\cite{DBLP:journals/corr/abs-2103-09504} will be referred to as predicted frames.
We found that 50000 epochs were sufficient to train the network.
Figure \ref{training_loss} show the convergence of the training loss, the validation Mean-Squared Error, (MSE) and the validation Learned Perceptual Image Patch Similarity (LPIPS) with the number of epochs.
A link to a Github repository that contains the trained network, as well as the list of optimal hyperparameter values, is included in the section {\it Methods}. 

\begin{figure}[H]
    \begin{subfigure}{.49\textwidth}
  \includegraphics[width=\textwidth]{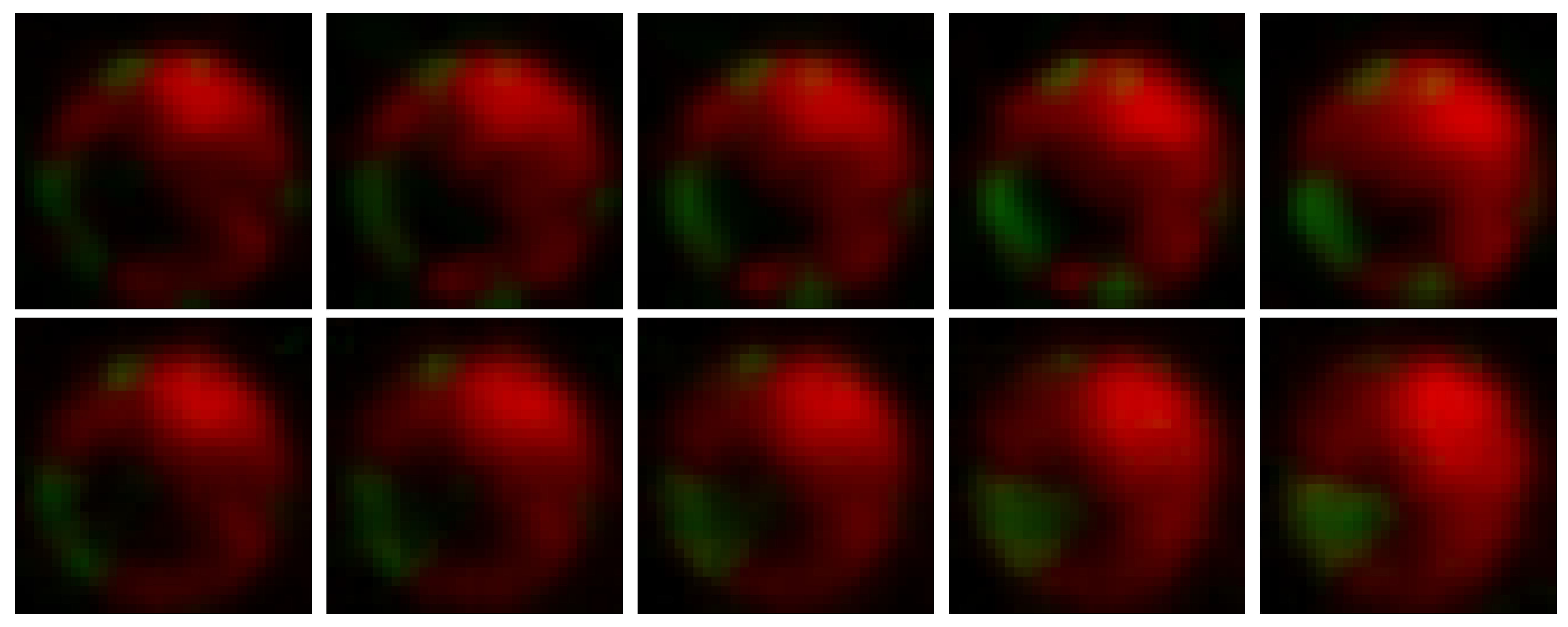}
  \subcaption{Test well 1}
  \end{subfigure}
    \begin{subfigure}{.49\textwidth}
  \includegraphics[width=\textwidth]{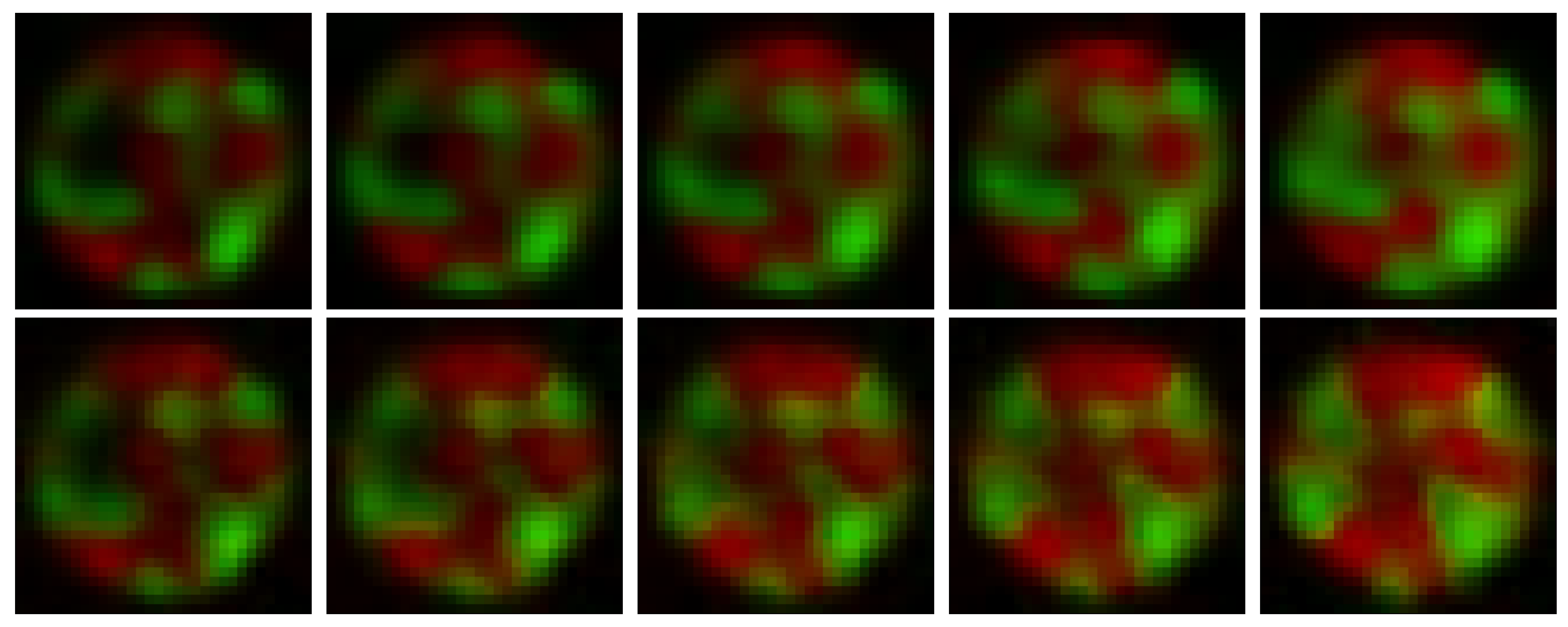}
  \subcaption{Test well 2}
  \end{subfigure}
    \begin{subfigure}{.49\textwidth}
  \includegraphics[width=\textwidth]{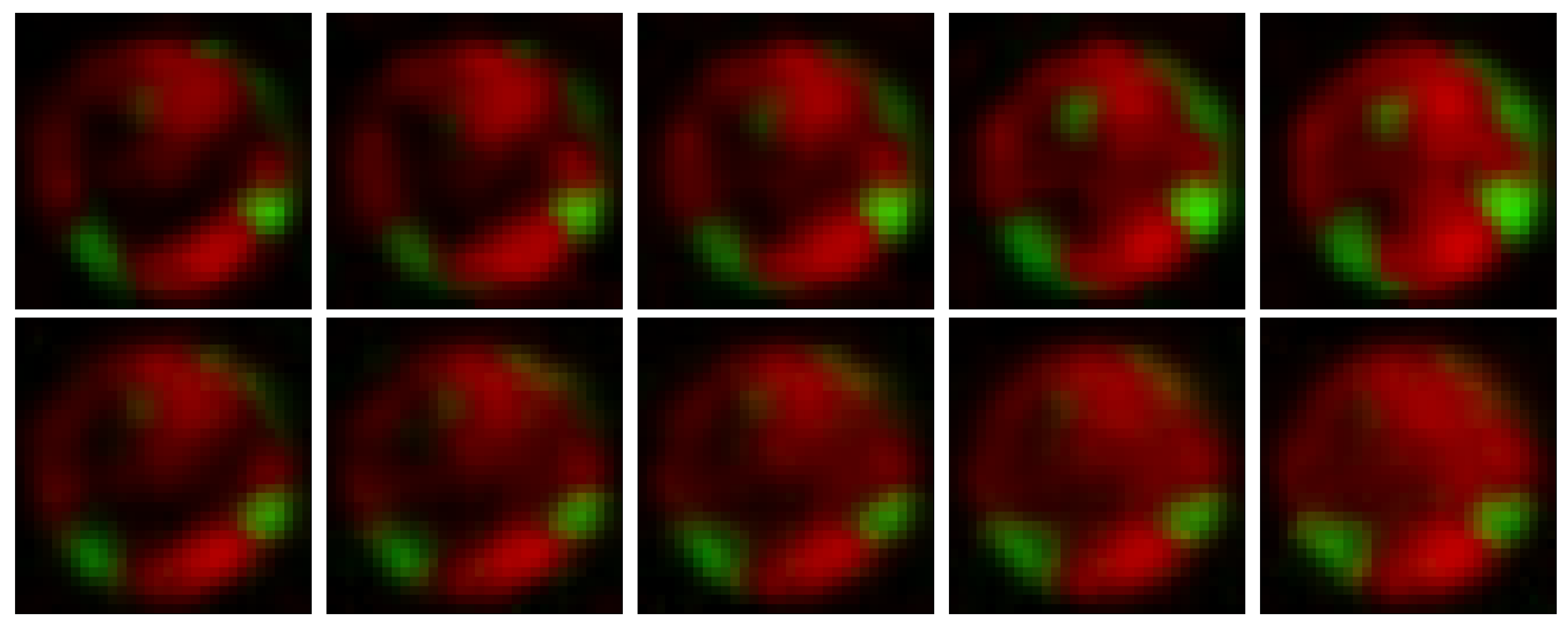}
  \subcaption{Test well 3}
  \end{subfigure}
    \begin{subfigure}{.49\textwidth}
  \includegraphics[width=\textwidth]{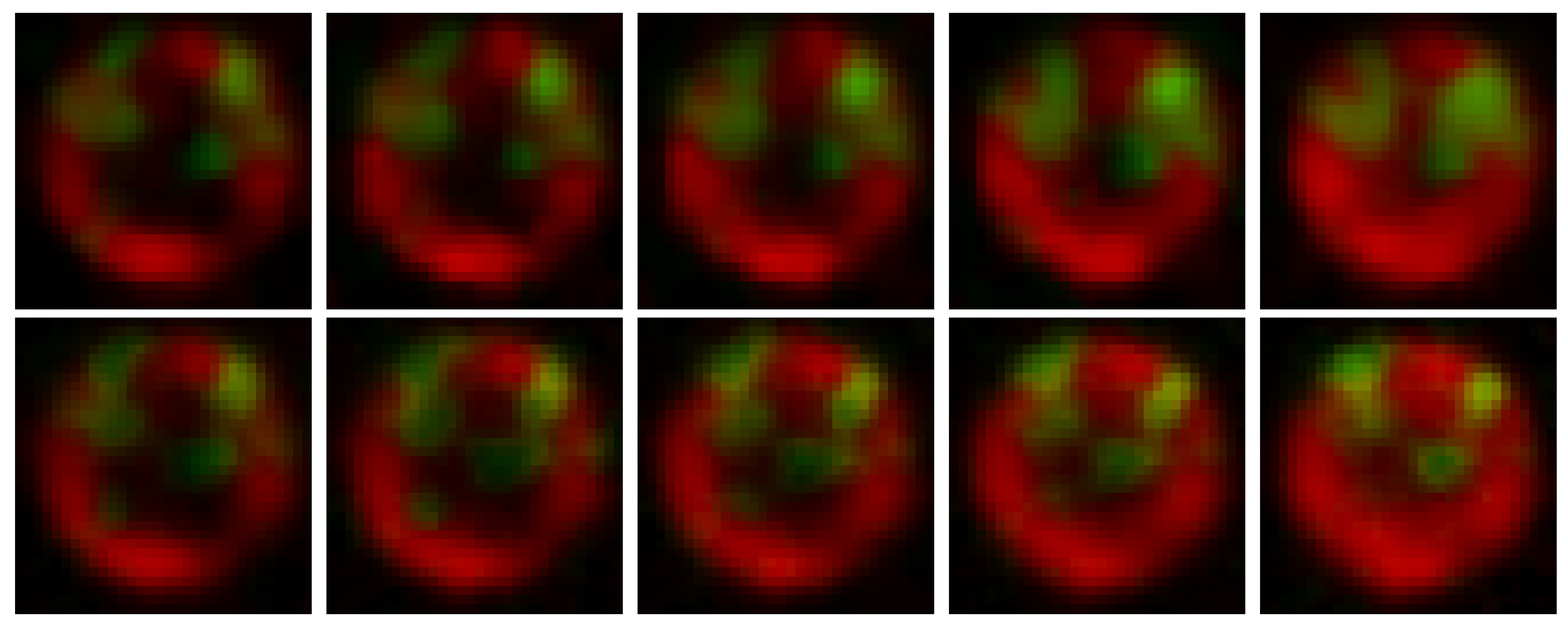}
  \subcaption{Test well 4}
  \end{subfigure}
  \caption{Qualitative comparison between the groundtruth and predicted frames for four wells in the test dataset. The upper panel represents the groundtruth frames, and the lower the predicted frames. The first image at the left of each panel represents the 11$^{th}$ frame, followed by 13$^{th}$, 15$^{th}$, 17$^{th}$ and 19$^{th}$. The wells are numbered according to their position in the test dataset.} 
  \label{bulk_comparisons}
\end{figure}

To assess the generalization and transfer ability of the trained network, we considered 4 different wells from the test dataset, and used PredRNN\cite{DBLP:journals/corr/abs-2103-09504} to predict the frames 11-20 for each well. The groundtruth and the corresponding predicted frames for each of these 4 wells are shown in Fig.\ref{bulk_comparisons}

\begin{figure}[H]
    \begin{subfigure}{.49\textwidth}
      \includegraphics[width=\textwidth]{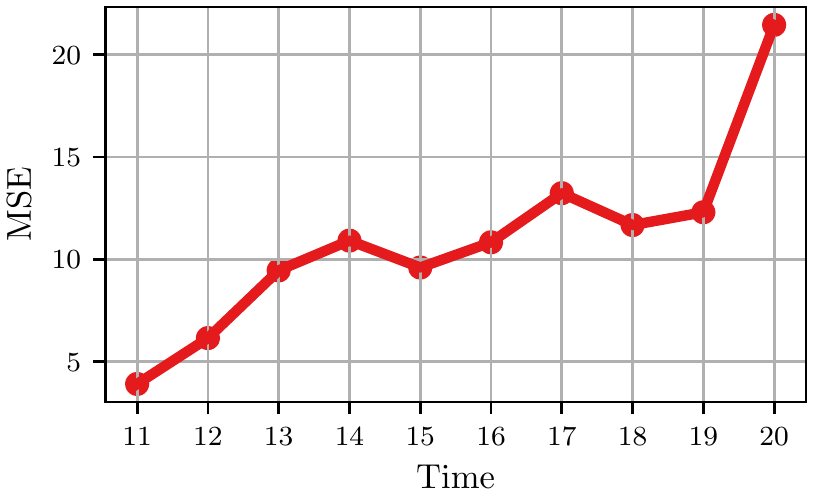}
      \subcaption{}
      \label{mse}
    \end{subfigure}
    \begin{subfigure}{.49\textwidth}
      \includegraphics[width=\textwidth]{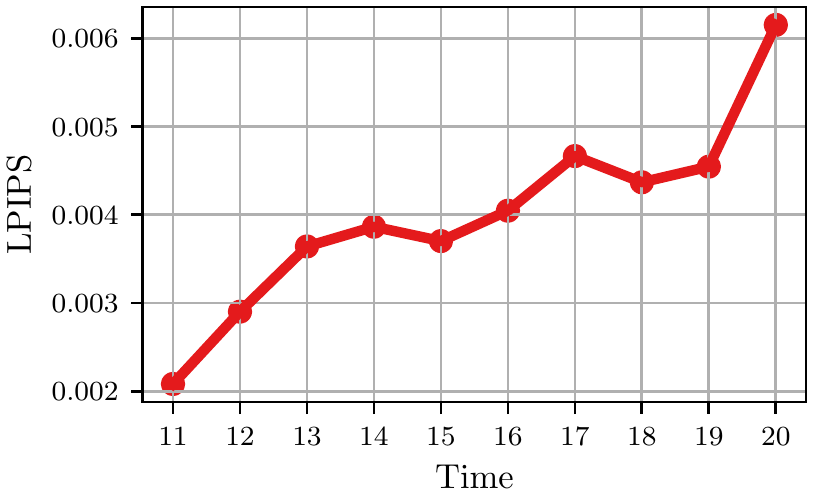}
      \subcaption{}
      \label{lpips}
    \end{subfigure}
  \caption{Quantitative comparison between the groundtruth and the predicted frames for the test wells in Fig.\ref{fig:fig2}. (a) average Mean-Squared Error (MSE) for all test wells; (b) average Learned Perceptual Image Patch Similarity (LPIPS) for all test wells. Each dot represents a frame.}
\end{figure}

 As seen in Fig.\ref{bulk_comparisons}, the qualitative  agreement between the groundtruth and predicted frames is very good. Quantitative agreement was assessed  by computing the per-frame and the average per-frame MSE and  LPIPS \cite{DBLP:journals/corr/abs-2103-09504} (these metrics compare the groundtruth and the predicted frames, and the smaller their values are, the more similar is the predicted frame to its corresponding groundtruth one pixelwise). Figures ~\ref{mse} and \ref{lpips} show the evolution of the per-frame MSE and LPIPS for all 4 test wells.
 Both MSE and LPIPS increase with time due to the accumulation of errors as prediction progresses (an expected result).
 Nonetheless, the  accuracy of both the short-term and long-term predictions is satisfactory and comparable to the results obtained in \cite{DBLP:journals/corr/abs-2103-09504} for the moving MNIST database.
 Furthermore, the average per-frame MSE and LPIPS  values obtained for the test set are 13.7 and 0.002, which is markedly smaller than the values 48.4 and 0.071 obtained in \cite{DBLP:journals/corr/abs-2103-09504} for the moving MNIST database.

Although MSE and LPIPS are useful for directly comparing images, they are not particularly informative for microbial experiments.
In this context, it is preferable to compare images with metrics that provide information on microbial growth.

 Timm {\it et al.} ~\cite{WOS:000407448100034} provided information  on microbial growth by measuring population growth curves.
 Green and red fluorescent proteins were used to tag the T6SS and non-T6SS mutants {\it \textit{P. aeruginosa}}, respectively, and the growth of each, as well as the growth of the community, was measured by plotting the intensity of the fluorescent signal \textit{vs.} time. 
Here we computed population curves by adding up, for each frame, the pixel values for green and red.

\begin{figure}[H]
    \centering
    \includegraphics[width=\textwidth]{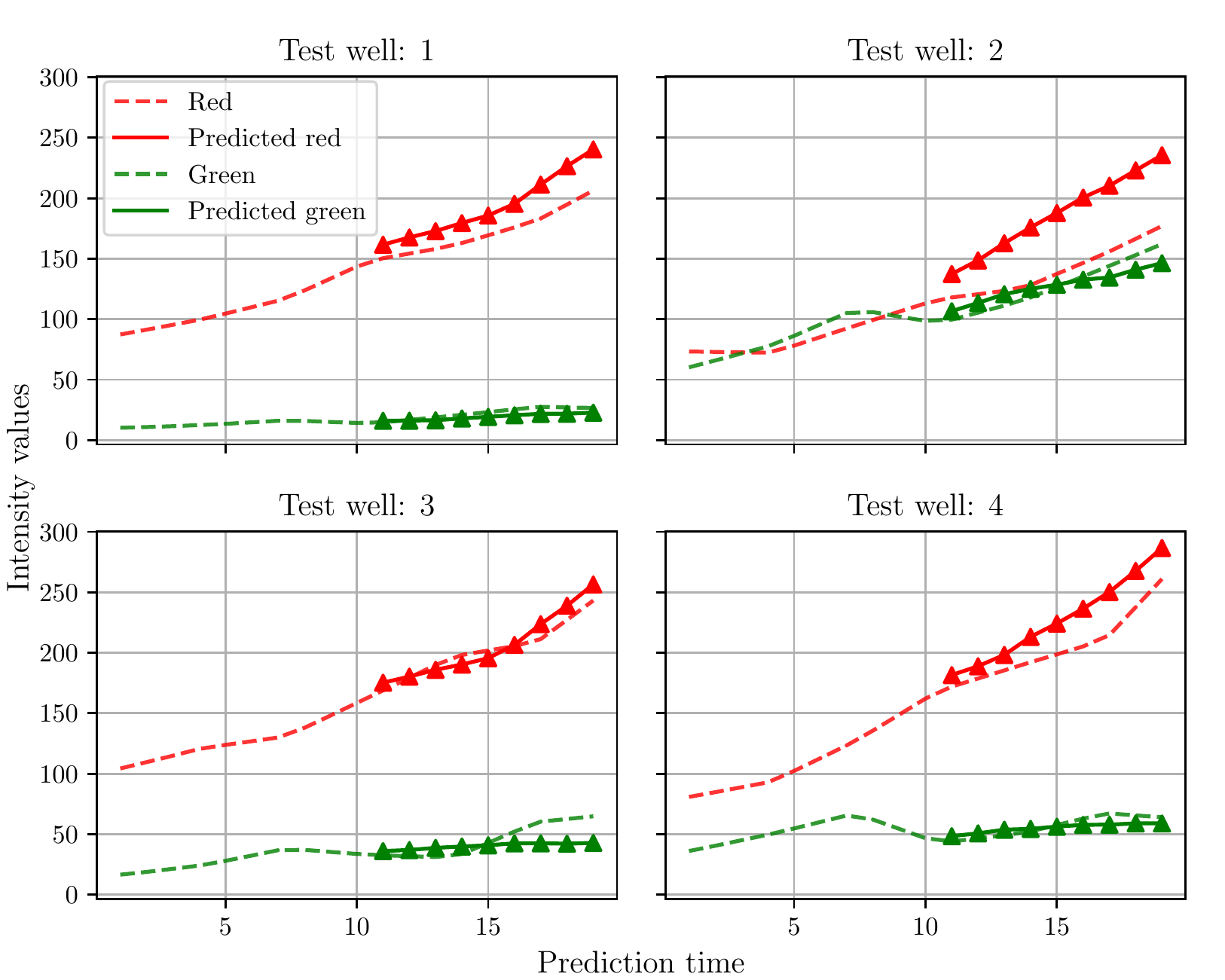}
    \caption{Population curve comparisons from groundtruth and PredRNN\cite{DBLP:journals/corr/abs-2103-09504} generated predictions}
    \label{global_population}
\end{figure}

Figure \ref{global_population} shows the population curves for the 4 test wells. As  time increases, the predicted  population curves differ more and more from the groundtruth ones.
This  is more apparent for the red curves (non-T6SS mutant type) than for the green (T6SS mutant type). The poorest comparison between the  groundtruth and the predicted population curves occurs for test well 2.

\begin{figure}[H]
  \includegraphics[width=\textwidth]{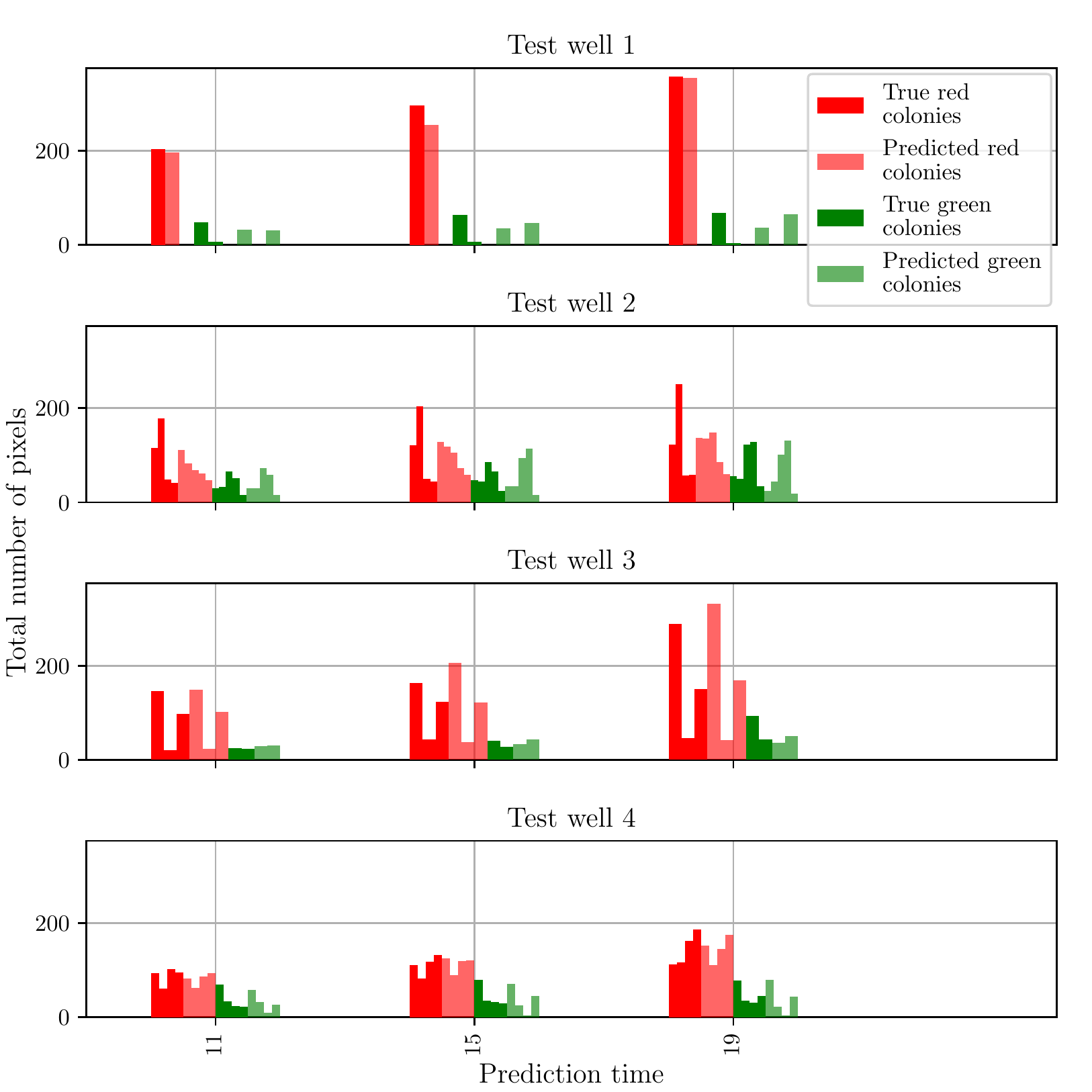}
  \caption{Comparison of individual colonies  between the groundtruth and predicted frames for wells in the test dataset. The wells are numbered according to their position in the microfluidics array.} 
  \label{colony_sizes}
\end{figure}

To inspect the population curves better and gain insights into the discrepancy observed for test well 2, we also counted the number of red and green colonies and their corresponding number of pixels.  The results for each of the test wells are shown in Fig. \ref{colony_sizes}.
For clarity, the data is shown for 3 different prediction times only. The agreement between prediction and groundtruth is in general very good, both in the number of colonies and their number of pixels. Some disagreement does exist in test well 1, where the number of predicted green colonies are overestimated from the beginning; in test well 4, where time steps 15 and 19 show one green colony less for the prediction sample; and more significantly, in test well 2, where the number of pixels for the red colonies are overestimated in the prediction.
In particular, the predictions produced the same number of red colonies than the groundtruth, but the predicted ones are larger than actually occurred. 

Interestingly, the disagreement between the predicted and groundtruth population curves was also more significant for test well 2 than for the other test wells.
We believe that the root of this disagreement stems from an imbalanced training data set.
Indeed, visual inspection of all 48 videos revealed that the majority of the wells contained a large red colony and a few smaller green ones.
Most likely, in simple terms,  PredRNN\cite{DBLP:journals/corr/abs-2103-09504} was trained to see more red than green, and this affects the predictions it makes.
 As a consequence, when wells with a balanced population of red and green mutants (see Fig.\ref{bulk_comparisons}) are encountered, red is overestimated, and  the difference between predictions and groundtruth are more significant.

\section{Conclusions}
We trained a RNN, called PredRNN\cite{DBLP:journals/corr/abs-2103-09504}, to perform video frame prediction of microbial growth for a population containing two mutants of {\it P. aeruginosa}.
For training, we used 48 videos that were generated with microfluidics and fluorescence microscopy. 
To asses the quality of the predictions, we used image-to-image metrics, in particular MSE and LPIPS, population curves and the distribution of individual colonies. 
Image-to-image metrics are commonly used in the machine learning community, whereas population curves and colony distribution are more pertinent to experiments in microbiology. 
It was found that PredRNN\cite{DBLP:journals/corr/abs-2103-09504} can, in general, predict correctly  the growth of this population. 
However, we also noticed that in some cases the predicted growth could be inaccurate.
In this particular setting, these inaccuracies were found to be caused by an imbalance training dataset, which happened to contain more examples wells with the  non-T6SS mutant growing faster than the T6SS-one. As a consequence, the network tended to favor the growth of  non-T6SS mutant over the other, and this lead to large errors in frame prediction for wells were both mutant happened to grow similarly.
This emphasizes the need to create large databases of microbial growth, which could be accomplished by combining microfluidics and fluorescence microscopy.
Thus, we believe that our results provide some of the first essential steps toward performing autonomous experiments in microbiology, were time spent in image acquisition could be saved by predicting growth with RNNs.
This could be accomplished by  deploying a trained PredRNN in an edge-computing device, on which only inference would be performed.
The inferred results could then be passed to a microscopy/microfluidics platform, and decisions, such as continuing to image a particular well or not, could then be made.

\section{Methods}\label{method}

The images used for training and prediction were collected in red and green autoflourescence corrected images of 48 sample wells.
Each well's image data consisted of 14 sequential 24 x 24 pixel images. 

To maximize the accuracy of the PredRNN\cite{DBLP:journals/corr/abs-2103-09504} training procedure, the sequences were adjusted to center the well in the image.
This was done using triangle thresholding and averaging of the sequence images in time to determine a bounding box on the brightest pixels.
After this adjustment, the samples were cut down to only the first 7 frames, when the bacterial growth is most dynamic and is growing to fill the well.
The subsequent frames 8 to 14 have somewhat stagnated in growth which presents a trivial case for prediction and were discarded.
The first 7 frames were subsequently augmented in time using linear interpolation across frames with the scikit-image Python package and within each image using bilinear interpolation from the OpenCV library.
Ultimately, this processing yielded a sequence of 20 frames of 32 x 32 pixel images for each well.

For improved training of the network, the images in all samples were transformed into Hue Saturation Value (HSV) image format.
This step represents an case-specific adjustment informed by our knowledge of the images.
More specifically, in the setting of two uniquely colored interacting populations (red and green), using the HSV format better separates measurements in each image channel as:
the presence of either population in a given pixel (value), the species of the population in the pixel (hue), and the concentration of the species in the pixel (saturation).
Using this colormap allowed us to use a smaller (and thus more easily trained) network while providing at least as much accuracy as can be attained using a Red-Green-Blue (RGB) image format and a larger network.

Once the original images had been centered, resized, and transformed, they were spatially augmented by flipping, rotation, blurring, and adding Gaussian noise using the AtomAI Python package~\cite{atomai}.
These transformations attempt to account for the bias introduced by the orientation, focus, and accuracy of the experimental imaging.
The temporal and spatial augmentation described above increased the sample size from 48 samples of 7 frames of 24 x 24 pixels to 392 samples of 20 frames of 32 x 32 pixels.
The 392 sample dataset was then randomly split into a training set (80\% of the data) and validation set (20\% of the data).
Several of the original wells (not included in the transformed data) were held back as a testing set. 

As is common with RNN architectures, there are a variety of network parameters that can be adjusted for improved results.
Due to the vast number of parameter permutations and the computational cost of each iteration, we focused our tuning to the layer size and learning rate of the models used in~\cite{DBLP:journals/corr/abs-2103-09504}.
This focused tuning was based on the original parameters used for weather prediction due to the similarity of the setting.
To evaluate the success of each set of model parameters, we used the mean-squared error (MSE), which is a pixel-wise comparison of images, and the learned perceptual image patch similarity (LPIPS).
The LPIPS metric was recently developed as an image similarity metric to mimic the perception of the human eye.
To do so, it compares the layer activations of a successfully trained image classification network, such as AlexNet, for the true image and the test image (in this case an RNN generated image).

By iterating across a range of layer sizes and learning weights, we ultimately determined to use layers of size 32 and a learning rate of $0.0003$.
After training for 50,000 iterations, the network converged in MSE and LPIPS in both training and testing sets as can be seen in Figure~\ref{training_loss}.

\begin{acknowledgement}
The machine learning modeling research was conducted at the Center for Nanophase Materials Sciences, which is a DOE Office of Science User Facility. C.R. thanks the SCGSR award for facilitating this work. The authors thank Scott Retterer and Andrea Timm for providing the experimental videos used for training PredRNN\cite{DBLP:journals/corr/abs-2103-09504}. All the simulations were run in CADES, https://cades.ornl.gov/. This material is also based upon work supported by the U.S. Department of Energy, Office of Science, Office of Workforce Development for Teachers and Scientists, Office of Science Graduate Student Research (SCGSR) program. The SCGSR program is administered by the Oak Ridge Institute for Science and Education (ORISE) for the DOE. ORISE is managed by ORAU under contract number DE‐SC0014664.
\end{acknowledgement}

\begin{suppinfo}
The videos, the script used to trained PredRNN\cite{DBLP:journals/corr/abs-2103-09504} and the trained model can be found in this github repository: https://github.com/miguel-fc/RNNmicrobial.git
\end{suppinfo}

\bibliography{references}

\end{document}